\documentclass{article}

\usepackage{microtype}
\usepackage{graphicx}
\usepackage{subcaption}
\usepackage{booktabs} 

\usepackage{hyperref}

\usepackage[preprint]{icml2026}

\usepackage{amsmath}
\usepackage{amssymb}
\usepackage{mathtools}
\usepackage{amsthm}
\usepackage{graphicx}

\usepackage[capitalize,noabbrev]{cleveref}

\theoremstyle{plain}

\theoremstyle{definition}

\theoremstyle{remark}

\usepackage[textsize=tiny]{todonotes}

\usepackage{xcolor}
\usepackage{amssymb}
\usepackage{pifont}
\usepackage{tabularx}
\usepackage{array}
\newcolumntype{Y}{>{\centering\arraybackslash}X}

\newcommand{\cmark}{\textcolor{green}{\checkmark}}
\newcommand{\xmark}{\textcolor{red}{\ding{55}}}   

\usepackage[most]{tcolorbox}
\tcbuselibrary{listings, breakable, skins}
\newtcolorbox{promptbox}[2][]{
  enhanced,
  breakable,
  colback=gray!2,
  colframe=gray!40!black,
  colbacktitle=gray!10,
  fonttitle=\bfseries,
  title={#2},
  sharp corners,
  boxrule=0.8pt,
  coltitle=black,
  #1
}
\usepackage{tcolorbox}
\usepackage{xcolor}
\usepackage{listings}
\usepackage{courier}
\usepackage[T1]{fontenc}

\lstdefinestyle{json}{
    basicstyle=\ttfamily\footnotesize,
    numbers=none,
    breaklines=true,
    showstringspaces=false,
    frame=single,
    rulecolor=\color{gray!40},
    backgroundcolor=\color{gray!5},
    keywordstyle=\color{blue!70!black},
    stringstyle=\color{teal!70!black},
}
\lstdefinestyle{python}{
    language=Python,
    basicstyle=\ttfamily\footnotesize,
    keywordstyle=\color{blue}\bfseries,
    stringstyle=\color{teal},
    commentstyle=\color{gray},
    breaklines=true,
    breakatwhitespace=true,
    showstringspaces=false,
    numbers=left,
    numberstyle=\tiny\color{gray},
    xleftmargin=2em,
    frame=none
}

\icmltitlerunning{World of Workflows}

\begin{document}

\twocolumn[
  \icmltitle{World of Workflows: A Benchmark for \\ Bringing World Models to Enterprise Systems}



  \icmlsetsymbol{equal}{*}
    \begin{icmlauthorlist}
        \icmlauthor{Lakshya Gupta*}{skyfall}
        \icmlauthor{Litao Li*}{skyfall}
        \icmlauthor{Yizhe Liu*}{skyfall}
        \icmlauthor{Sriram Ganapathi Subramanian}{skyfall}
        \icmlauthor{Kaheer Suleman}{skyfall}
        \icmlauthor{Zichen Zhang}{skyfall}
        \icmlauthor{Haoye Lu}{skyfall}
        \icmlauthor{Sumit Pasupalak}{skyfall}
\end{icmlauthorlist}

\icmlaffiliation{skyfall}{Skyfall AI}

\icmlcorrespondingauthor{Litao Li}{mike@skyfall.ai}



\vskip 0.3in




]



\printAffiliationsAndNotice{\icmlEqualContribution}

\begin{abstract}
Frontier large language models (LLMs) excel as autonomous agents in many domains, yet they remain untested in complex enterprise systems where hidden workflows create cascading effects across interconnected databases. Existing enterprise benchmarks evaluate surface-level agentic task completion similar to general consumer benchmarks, ignoring true challenges in enterprises, such as limited observability, large database state, and hidden workflows with cascading side effects. We introduce \textbf{W}orld \textbf{o}f \textbf{W}orkflows (WoW), a realistic ServiceNow-based environment incorporating 4,000+ business rules and 55 active workflows embedded in the system, alongside WoW-bench, a benchmark of 234 tasks evaluating constrained agentic task completion and enterprise dynamics modeling capabilities. We reveal two major takeaways: (1) Frontier LLMs suffer from dynamics blindness, consistently failing to predict the invisible, cascading side effects of their actions, which leads to silent constraint violations, and (2) reliability in opaque systems requires grounded world modeling, where agents must mentally simulate hidden state transitions to bridge the observability gap when high-fidelity feedback is unavailable. For reliable and useful enterprise agents, WoW motivates a new paradigm to explicitly learn system dynamics. We release our GitHub~\footnote{\href{https://github.com/Skyfall-Research/world-of-workflows}{GitHub}} for setting up and evaluating WoW.
\end{abstract}

\section{Introduction}

Enterprise systems contain relational tables, workflow engines, business rules, and application modules that interact through complex, often opaque, chains of data flow. These systems form an intractably large and partially observable state space -- one where a single user action can trigger multiple cascading updates across dependent tables~\citep{wei2005understanding, lairet2024unavoidable}.

Despite impressive progress in general-domain autonomy, frontier LLMs remain unreliable in enterprise environments. Recent enterprise benchmarks~\citep{boisvert2024workarena++, huang2025crmarena, dai2025scuba, levy2024st} indicate low task success rates. Such benchmarks reveal several dimensions of challenges including hallucinations, domain understanding, UI navigation, and goal following. However, these challenges remain on the same surface as general consumer benchmarks with straightforward tasks and little constraint satisfaction requirements. In other words, they do not expose the true challenges of enterprise systems, including limited observability with intractably large database state, workflows with side effects, and multi-hop data dependencies that govern real enterprise behavior. To address these limitations, we introduce WoW, a realistic and workflow-centric ServiceNow-based~\footnote{\href{https://developer.servicenow.com}{ServiceNow Developer Instance}} enterprise system containing over 4,000 business rules and 55 active workflows. 
We also build WoW-bench, the first enterprise benchmark designed to evaluate LLMs both as enterprise world models and agents. 
WoW-bench provides 234 evaluation tasks that explicitly incorporate workflow effects into four task categories: autonomous task completion, data-level constraint understanding, dynamics prediction, and tool prediction.
We use WoW-bench to answer several critical research questions regarding the practicality and usefulness of enterprise agents. 

\begin{figure*}[ht]
    \centering
    \includegraphics[width=0.9\textwidth]{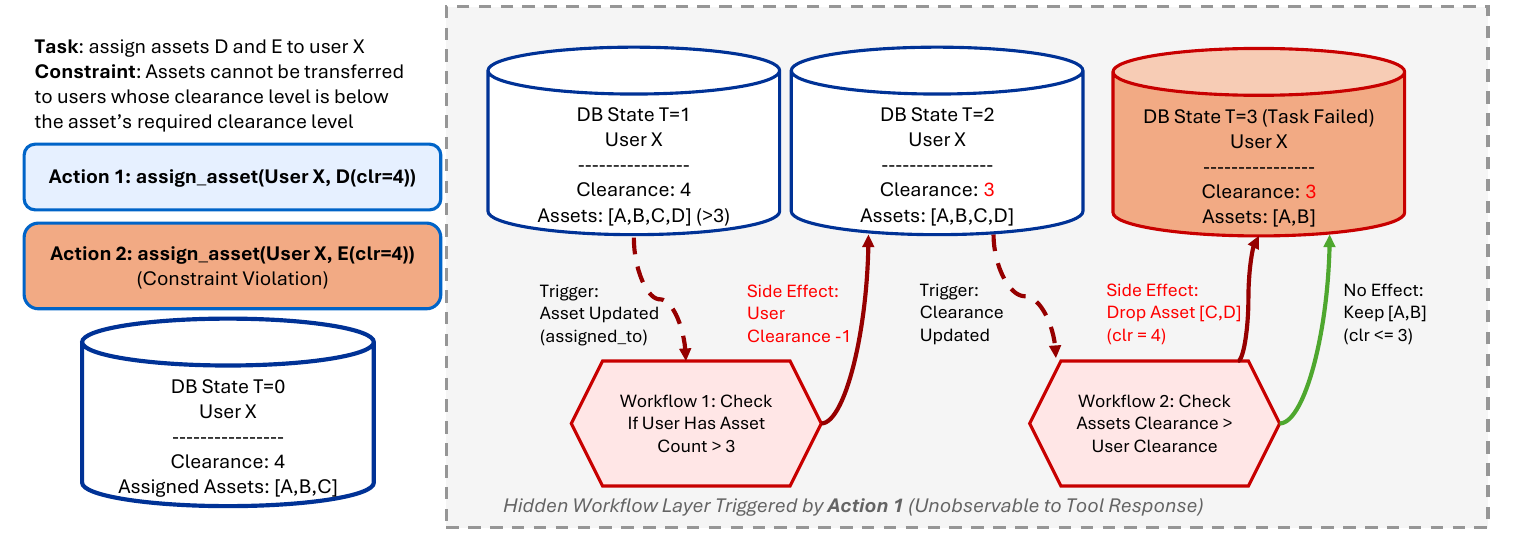}    
    \caption{Illustration of Cascading Workflow Failures in WoW. This diagram demonstrates how a single agent action can trigger a sequence of hidden state changes that violate constraints.}
    \label{fig:wf_example}
\end{figure*}

Firstly, how do hidden workflows dictate an agent’s ability to complete tasks and satisfy constraints? 
To answer this, we need to carefully design realistic workflows invisible to the agents and embed them in the system. We give an example in~\Cref{fig:wf_example} of how workflows lead to task failure and constraint violation. When agents observe limited information from tool responses, which primarily contain information of the tool execution status, task failure and constraint violation remain unexposed. From the agent's perspective in this example, assigning assets D and E to User X is compliant, as their clearance levels are both equal to that of user X.

Secondly, to what extent does bridging the "dynamics gap" with oracle-level state visibility unlock reliability in long-horizon tasks, and where do fundamental reasoning failures persist despite perfect information? 
WoW provides two forms of observations as an ablation study: tool response, which provides direct API feedback, and table audit logs, a structured representation of database state changes. Tool responses or screenshots abstract away the state change and hide the environment dynamics. By exposing the database state changes through audit logs, we aim to study how important it is to provide the necessary "evidence" for agents to trace side effects. Through our experiment, using audit logs as observation increases the task success rate by at most 7x, indicating the necessity of state visibility. 

Lastly, can frontier LLMs effectively function as zero-shot world models capable of predicting the cascading side effects of dependent actions and tracking symbolic state changes? Though audit logs provide significant uplift to task completion, assuming the availability of such information is not realistic in many enterprise environments, as they incur high cost and latency~\citep{microsoft2025auditing} and often require escalated access~\citep{nist2017intro}. Through experiments, we evaluate the agents' capability of precise state tracking and tool use, which plays a critical role in real-world environments when oracle state is not visible.

our contributions include the following: 
\begin{itemize}
    \item We introduce WoW, a high-fidelity enterprise environment that simulates a realistic, workflow-centric, and complex system.
    \item We present WoW-bench, a novel benchmark designed to evaluate frontier LLMs' capabilities to reason, predict, and operate in an agentic and world modeling setting for completing constrained tasks and understanding system dynamics.
    \item We provide extensive evaluation showing that frontier LLMs struggle to track database and entity state changes, predict workflow effects, and follow constraints in long-horizon tasks. We motivate the need for a new agent learning paradigm to incorporate system dynamics modeling.
\end{itemize}

\section{Related Work}
\subsection{World Models}
In embodiment and robotics, world models serve as predictive simulators for physical tasks~\citep{assran2025v, ha2018world, ding2025understanding}. However, their reliance on continuous dynamics and dense sensor measurements does not transfer to the symbolic, discrete, and workflow-driven nature of enterprise systems.
Text-based world models examine whether LLMs can learn symbolic environment dynamics expressed in natural language. TextWorld~\citep{cote2018textworld}, TextWorld Commonsense~\citep{murugesan2021text}, and STARLING~\citep{basavatia2024starling} treat the world as a set of textual predicates and model local transitions. Such worlds lack realism with tiny state space and simple dynamics.
MAPS~\citep{aroca2025mini} and VendingBench~\citep{backlund2025vending} benchmark LLMs to anticipate consequences under uncertainty and strategically operate a complex business. 
In these environments, world models operate on compact and well-structured state representations and controllable environment dynamics. In contrast, enterprise systems exhibit limited observability and workflow-driven database state updates. This gap motivates the need for world-model benchmarks operating over symbolic, relational, workflow-governed environments.

\begin{table*}[ht]
\centering
\caption{Comparison matrix between WoW-bench and previous benchmarks. WoW focuses on incorporating and evaluating world modeling, constraint understanding, and task completion in realistic enterprise environment with underlying workflows.}
\resizebox{\textwidth}{!}{
\begin{tabular}{cccccc}
\toprule
\textbf{Benchmarks} & Realistic Environment & Enterprise Tasks & Constraint Following & World Model Evaluation & Complex Workflows  \\ \midrule
WorkArena++~\citep{boisvert2024workarena++} & \cmark & \cmark & \xmark & \xmark & \xmark \\ 
CRMArena-Pro~\citep{huang2025crmarena} & \cmark & \cmark & \xmark & \xmark & \xmark \\ 
SCUBA~\citep{dai2025scuba} & \cmark & \cmark & \xmark & \xmark & \xmark \\ 
ST-WebAgentBench~\citep{levy2024st} & \cmark & \cmark & \cmark & \xmark & \xmark \\ 
WorkBench~\citep{styles2024workbench} & \xmark & \cmark & \xmark & \xmark & \xmark \\
$\tau^2$-bench~\citep{barres2025tau} & \xmark & \xmark & \xmark & \xmark & \xmark \\
MCPToolBench++~\citep{fan2025mcptoolbench++} & \cmark & \xmark & \xmark & \xmark & \xmark \\
MCP-Universe~\citep{luo2025mcp} & \cmark & \xmark & \xmark & \xmark & \xmark \\
WoW-bench & \cmark & \cmark & \cmark & \cmark & \cmark \\
\bottomrule
\end{tabular}
}%
\label{tab:bench_comparison}
\end{table*}

\subsection{General Agentic Benchmarks}
LLM-based agents in interactive environments have been extensively studied to perform different types of tasks, such as software engineering~\citep{chen2021evaluating, austin2021program, jain2024livecodebench}, open-ended computer use~\citep{yao2022webshop, zhou2023webarena, xie2024osworld}, and tool use~\citep{liu2023agentbench, wang2025software}. 
Since the primary interaction method in WoW is MCP tools, we will not discuss the details of general software engineering and web agents. Tool calling has become a core capability for LLMs, as they need to decide when and how to use the correct tools given specific tasks in natural language~\citep{mohammadi2025evaluation}. Several benchmarks evaluate the tool use accuracy~\citep{huang2023metatool, patilberkeley}, where tools are manually curated by the developers, and have simple functionalities and little realism. Others evaluate MCP tools~\citep{luo2025mcp, wang2025mcp, fan2025mcptoolbench++, wu2509mcpmark}, which provide a wide array of real-world applications, such as finance, file system, maps, and search. Although the aforementioned benchmarks cover a wide range of domains, they are likely seen during the large-scale training of LLMs, making the usefulness in extremely domain-specific environments untested. Enterprises are heavily domain-engineered with system-specific requirements and unique interactions. 

\subsection{Enterprise Benchmarks}
Enterprise benchmarks can be categorized into UI-based and API-based. UI-based benchmarks require visual reasoning and action grounding. 
WorkArena++~\citep{boisvert2024workarena++} provides both atomic and compositional tasks that require problem solving and logical reasoning in ServiceNow developer instance. 
SCUBA~\citep{dai2025scuba} tests a range of agent capabilities including UI navigation, data manipulation, workflow automation, information retrieval, and troubleshooting. 
ST-WebAgentBench~\citep{levy2024st} extends the focus on agent study from task completion to safety and trustworthiness. 
Wonderbread~\citep{wornow2024wonderbread} investigates the ability of agents to document business processes and apply them in real tasks. 
UI-based agents rarely represent real-world use cases, as they introduce long trajectories and compounding errors compared to efficient interaction through APIs. 
API-based benchmarks such as CRMArena-Pro~\citep{huang2025crmarena} uses query and knowledge base search tools for CRM-related tasks, but only involves read-only operations, which limits evaluation for end-to-end autonomous task completion. WorkBench~\citep{styles2024workbench} uses a self-hosted sandbox environment without realistic enterprise system, indicating limited practicality. 
The largest limitation for the above benchmarks is that their tasks still evaluate the same aspects as consumer and general agentic tasks, overlooking the distinct properties in enterprise environments such as complex data flow and unobvious constraint satisfaction. WoW fills the gap with large database state, heavy workflows, and carefully designed constraints.

\begin{figure*}[ht]
    \centering
    \includegraphics[width=0.9\textwidth]{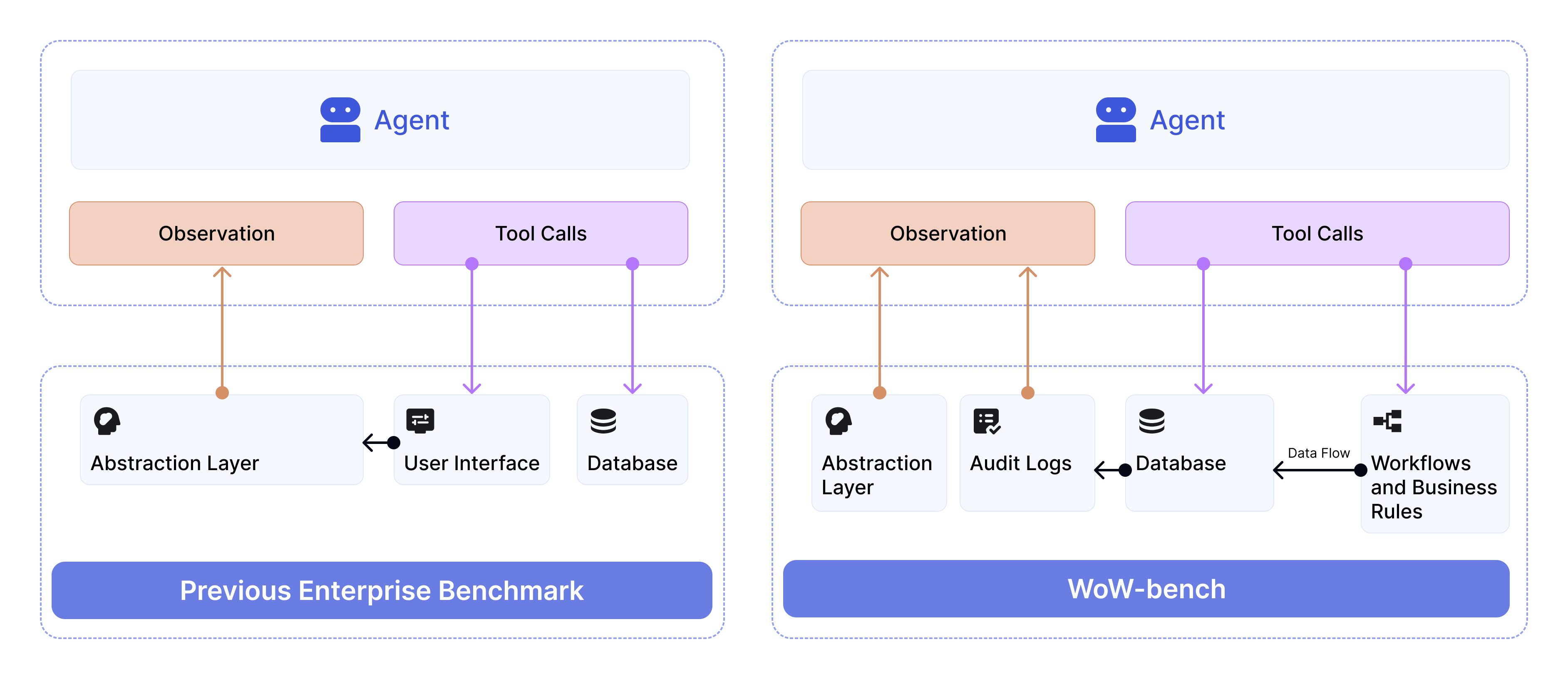}    
    \caption{WoW vs. previous enterprise benchmarks. WoW includes complex workflows and business rules, which introduce hidden data flow and API calls that can impact the database state. Compared to previous environments, which do not involve workflows, WoW requires agents to model the system dynamics to correctly carry out tasks and follow constraints. We introduce an augmentation to the observation space with table audits as tractable state changes to provide low-level information to agents.}
    \label{fig:wow_framework}
\end{figure*}

\section{WoW Environment}
WoW is an enterprise environment featuring realistic IT functionalities, covering sub-domains across multiple management systems including \textbf{user}, \textbf{incident}, \textbf{asset}, \textbf{knowledge base}, \textbf{catalog}, and \textbf{expense}.
An agent interacts with the environment through MCP tools as actions, and perceives the partial observation of the world via tool response or table audits. 
WoW is built using the ServiceNow developer instance, a free sandbox platform to experiment and test custom applications. Upon initialization, mock data are automatically populated. ServiceNow allows flexible and customizable controllability as an environment, including action, observability, and workflow definitions.

\subsection{Environment Specification}
WoW is a partially observable environment, and we model the tasks in WoW as an Partially Observable Markov Decision Process (POMDP). It is formally defined by the tuple $\mathcal{M} = (\mathcal{U}, \mathcal{S}, \mathcal{A}, \mathcal{T}, \mathcal{O}, \Omega)$. $\mathcal{U}$ is the user query consisting of task and constraint descriptions. $\mathcal{S}$ is the state space, which is the entire database state. $\mathcal{A}$ is the action space. $\mathcal{T}:\mathcal{S} \times \mathcal{A} \rightarrow S'$ is the state transition function. $\mathcal{O}$ is the observation space. We study two different observation functions: $\Omega_{tool}:\mathcal{S} \times \mathcal{A} \rightarrow \mathcal{O}_{tool}$ and $\Omega_{audit}:\mathcal{S} \times \mathcal{A} \rightarrow \mathcal{O}_{audit}$, which provide two choices of observations: $\mathcal{O}_{tool}$ for tool response and $\mathcal{O}_{audit}$ for oracle table audit logs.

\textbf{Action} 
Let $\mathcal{A}$ denote the discrete action space of the environment. An action $a_t \in \mathcal{A}$ at time step $t$ is defined as a tuple consisting of a selected MCP tool and its associated parameters:

\begin{equation}
a_t = ( \eta, \mathcal{P} )
\end{equation}

where $\eta \in \mathcal{N}$ represents the unique \textbf{tool name} from the set of available tools. $\mathcal{P} = \{ (k_1, v_1), \dots, (k_m, v_m) \}$ is the set of $m$ \textbf{parameter} key-value pairs required by the tool. An example of tool specification can be found in the Appendix~\ref{appendix:tool}. 
The value $v_j$ for any parameter $k_j$ is drawn from a heterogeneous domain that is defined as the union of four distinct data types: categorical options, numerical values, structured date and time, and free-form text strings. $\mathcal{A}$ is combinatorially large due to free-form parameters, and also requires high precision for semantically correct tool execution. The causal effects of actions are difficult for agents to predict in a zero-shot enterprise setting, compared to general consumer settings.

\textbf{State Space} Let $\mathcal{S}$ denote the global state space of the enterprise environment. The state $s_t \in \mathcal{S}$ at time $t$ represents the complete configuration of the underlying relational database. Formally, $s_t$ is the union of all data entries across all tables:

\begin{equation}
s_t = \bigcup_{i=1}^{N} \mathcal{R}_i^{(t)}
\end{equation}

where $\mathcal{R}_i^{(t)}$ represents the set of all records and their field values in the $i$-th table at time $t$.
\newline
\textbf{Intractability:} In realistic enterprise settings, $N$ is large (e.g., thousands of tables) and the number of records is vast. Consequently, the full state $s_t$ is \textbf{intractably large} and never fully observable to the agent. This effect is magnified in a workflow-centric setting, where the state transition function is more complex and involves hidden mechanics of the system. The agent must rely on partial observations to infer the state.

\textbf{Observation Space} WoW is a partially observable environment, representing a real-world case that aligns with the big world hypothesis~\citep{javed2024big}, which states that the true world state is intractably large. The observation space $\mathcal{O}$ determines the feedback the agent receives after executing action $a_t$. We define two distinct observation settings to evaluate the agent's ability to model system dynamics.

\paragraph{1. Standard Observation (Tool Response)}
In the standard setting, the observation $o_t^{\text{tool}} \in \mathcal{O}_{tool}$ consists solely of the immediate output returned by the tool execution (e.g., API success messages, error codes, or fetched record data)~\citep{wang2025mcp, fan2025mcptoolbench++}.

\begin{equation}
o_t^{\text{tool}} = \text{Resp}(a_t, s_{t+1})
\end{equation}

where $\text{Resp}(\cdot)$ is the string or JSON output defined by the tool's interface. This observation mode hides the side effects and downstream cascading changes triggered by hidden workflows, leading to significant limitation of observability and decision making capability.

\paragraph{2. Oracle Observation (Table Audits)}
To investigate whether providing visibility into workflow effects aids task completion, we introduce an oracle observation setting. This setting augments the tool response with a set of \textit{Table Audits} $\Delta_t$, representing the differential state changes:

\begin{equation}
o_t^{\text{audit}} = ( o_t^{\text{tool}}, \Delta_t )
\end{equation}

Here, $\Delta_t$ is the set of all discrete database updates that occurred during the transition from $s_t$ to $s_{t+1}$ (including those caused by hidden asynchronous workflows). Each element $\delta \in \Delta_t$ is a structured tuple:

\begin{equation}
\delta = (\text{TableName}, \text{ColumnName}, \text{OldValue}, \text{NewValue})
\end{equation}

This ablation allows us to measure the performance gap between agents operating on surface-level feedback versus those with grounded visibility into the workflow effects. Note that audits are still the products of system dynamics, where the true specifications of workflows are not exposed to agents, making the setting remain partially observable.

\subsection{Workflows}
There are 55 active workflows and 4.8K business rules in ServiceNow. They represent business applications written by human programatically that are triggered when criteria are met~\citep{servicenow_flow_designer_vs_business_rules}. Business rules provide immediate, atomic, and database-driven logic to perform actions, such as setting the value of one column based on the rules of other values in the same table. Workflows, on the other hand, provide process orchestration, rule implementation, calculation, and multi-step and multi-system interactions. We illustrate that workflows can be very complex, where some examples are shown in~\Cref{fig:workflow_screenshot}. They can be defined as full workflows or modular sub-flows. We show a detailed example in~\Cref{fig:wf_example}, that it is possible for workflows to trigger in a cascading fashion. Consider two workflows: 
\begin{enumerate}
    \item \textbf{WF1: User Clearance Decrement}: triggers on asset assignment changes; checks if the user being assigned to has more than 3 assets; results in the user clearance level decrement by 1.
    \item \textbf{WF2: Unassign Asset for Asset Clearance Compliance}: triggers on user clearance level changes; checks if any assigned assets have clearance levels exceeding that of the user; results in removal of such assets from the user.
\end{enumerate}
The agent attempts to assign two assets to User X. While the first assignment is locally valid, it triggers WF1, which downgrades the user's clearance. This state change subsequently triggers WF2, creating a constraint violation for existing assets. The agent, observing only successful API responses, remains unaware of the downstream failure.

\section{WoW-bench}
We argue that although environments in previous benchmarks are partially observable, their tasks are not designed to utilize such property. In most cases, such tasks can be completed solely based on observed information, effectively turning them into a fully observable setting.
To address the limitations of lack of hidden workflows, we present WoW-bench, a challenging and hybrid benchmark designed to evaluate LLM agents on system dynamics understanding, constraint satisfaction, and agentic tasks with cascading effects. We investigate enterprise agents from a new perspective: can agents perform and plan under complex data flow and hidden interactions, resulting in potentially unexpected side effects? To answer this, WoW-bench contains 234 tasks in total, split into four task categories: 67 action prediction tasks, 67 audit prediction tasks, 50 constraint understanding tasks, and 50 agentic tasks. We create cleanup functions for each task template that searches and deletes all the task description specific data from the instance. The experiments can be reproduced with different models and hyperparameters as many times as needed. 
%


\subsection{Task Categories}
We structure the benchmark into four distinct tasks, each designed to answer a specific research question regarding the agent's ability to function as a world model. The details of annotation process can be found in~\Cref{appendix:dataset}.

\subsubsection{Constraint Understanding}
\textbf{Research Question}: How do hidden workflows dictate an agent's ability to complete tasks and satisfy constraints?

\textbf{Rationale}: In standard benchmarks, constraints are usually explicit and static (e.g., "do not spend more than \$50"). In WoW, constraints are dynamic and often violated silently by the system itself (e.g., a background workflow stripping a user's clearance after an asset assignment). We posit that agents fail not because they ignore instructions, but because they cannot trace the causal link between their action and the system's hidden reaction.

\textbf{Design}: We curated 50 evaluation trajectories that explicitly trigger hidden workflows to create silent constraint violations. These tasks are designed to be realistic traps: the agent performs a locally valid action that triggers a remote, unobservable violation (e.g., Flagged articles should not be published). This category isolates the agent's ability to detect valid-but-unsafe state transitions driven by the environment's hidden logic. 

\subsubsection{Agentic Task Completion}
\textbf{Research Question}: To what extent does bridging the "dynamics gap" with oracle-level state visibility unlock reliability in long-horizon tasks?

\textbf{Rationale}: A key hypothesis of this work is that the "unreliability" of current agents stems from an observability gap. Tool responses ($\mathcal{O}_{tool}$) abstract away the side effects that define enterprise safety. By comparing performance against an oracle observation space ($\mathcal{O}_{audit}$), we can determine if failures are due to a lack of reasoning capability or simply a lack of information.

\textbf{Design}: We designed 50 long-horizon agentic tasks (avg. 13 steps) that require multi-hop symbolic reasoning. For example, to safely assign a role, the agent must check the user's manager and ensure no conflict exists—data points spanning three different tables. Crucially, we conduct an ablation study on these tasks using both $\mathcal{O}_{tool}$ and $\mathcal{O}_{audit}$ to quantify how much gain oracle state differences improve reliability and where reasoning failures persist even with perfect information.

\subsubsection{Action and Audit Prediction}
\textbf{Research Question}: Can frontier LLMs effectively function as zero-shot world models capable of predicting the cascading side effects of dependent actions?

\textbf{Rationale}: Since high-fidelity audit logs are often unavailable in real deployments due to cost or security, reliable agents must eventually possess an internal world model—the ability to mentally simulate the invisible physics of the enterprise.

\textbf{Design}: We isolate this simulation capability through Audit Prediction (Forward Dynamics) and Action Prediction (Inverse Dynamics) tasks. Unlike standard tool-use benchmarks that use random samples, we employ a Tool-Dependency Graph Sampling technique (see~\Cref{appendix:dataset}). This constructs valid, connected trajectories where tool outputs feed into subsequent inputs, ensuring agents are evaluated on realistic, multi-hop data flows rather than disjoint actions. This measures the agent's raw ability to predict state changes without the noise of planning or goal tracking.


\begin{table*}[ht]
\centering
\caption{Evaluation result for agentic task completion. TSR is task success rate, and TSRUC is task success rate under constraint. Average cost per task completion is also calculated.}
\begin{tabularx}{\textwidth}{c Y Y Y Y Y Y}
\toprule
\textbf{Metric} & TSR w/$\mathcal{O}_{audit}$   & TSRUC w/$\mathcal{O}_{audit}$ & Cost/Task w/$\mathcal{O}_{audit}$ & TSR w/$\mathcal{O}_{tool}$  & TSRUC w/$\mathcal{O}_{tool}$ & Cost/Task w/$\mathcal{O}_{tool}$\\ \midrule
GPT-5.1 &  32\% & 14\% & \$0.41 & 22\% &  2\% & \$0.14\\
Gemini-3-Pro & 42\% &  16\% &  \$0.64 &  38\% & 6\% &  \$0.41\\
Sonnet-4.5 & 58\% & 30\% & \$3.60 &  32\%  & 4\% & \$1.79\\
Opus-4.5 & 36\% & 14\% & \$5.00 &  26\%  & 8\% & \$3.08\\ \bottomrule
\end{tabularx}

\label{tab:agentic}
\end{table*}

\subsection{Evaluation Metrics}

We define quantitative metrics to measure agent reliability and world modeling capability. 

\subsubsection{Constraint \& Task Reliability}
For agentic tasks and constraint understanding, we evaluate whether agents can achieve goals without triggering hidden violations. Let $\mathcal{U}$ be the set of tasks. For each task $u \in \mathcal{U}$, we define two binary functions based on the final database state $S_{final}$:

\begin{itemize}
    \item $G(u,S_{final})$: Returns $1$ if the task goal is satisfied (or impossibility is correctly identified), else $0$.
    \item $V(u,S_{final})$: Returns $1$ if any constraint was violated during execution, else $0$.
\end{itemize}

We report \textbf{Task Success Rate (TSR)} and \textbf{Task Success Rate Under Constraint (TSRUC)}:

\begin{align}
TSR &= \frac{1}{|\mathcal{U}|} \sum_{u \in \mathcal{U}} G(u,S_{final}) \\
TSRUC &= \frac{1}{|\mathcal{U}|} \sum_{u \in \mathcal{U}} G(u,S_{final}) \times (1 - V(u,S_{final}))
\end{align}

For \textbf{Constraint Understanding} tasks specifically, success requires identifying the exact constraint violated and the action responsible (Exact Match). In addition, we also include a cost metric in US dollars for average task completion.

\subsubsection{Dynamics Modeling}
To evaluate the agent as a world model, we measure its ability to predict state transitions ($s_t \xrightarrow{a_t} s_{t+1}$).

\paragraph{Audit Prediction (Forward Dynamics):}
We measure the accuracy of predicted state changes (table audits) using \textbf{Intersection over Union (IoU)}. A match requires exact equality of the tuple \texttt{(Table, Column, OldVal, NewVal)}.

\begin{equation}
IoU_{Audit} = \frac{1}{K} \sum_{t=1}^{K} \frac{|\hat{\mathcal{O}}_{audit}^{(t)} \cap \mathcal{O}_{audit}^{*(t)}|}{|\hat{\mathcal{O}}_{audit}^{(t)} \cup \mathcal{O}_{audit}^{*(t)}|}
\end{equation}

\paragraph{Action Prediction (Inverse Dynamics):}
We evaluate the agent's ability to infer the action $a_t$ that caused a state change. We report \textbf{Tool Name Accuracy} ($ACC_{Type}$) and \textbf{Full Action Accuracy} ($ACC_{Full}$), where the latter requires an exact match of the tool name and all parameter values.

\begin{figure*}[ht]
    \centering
    \includegraphics[width=0.9\textwidth]{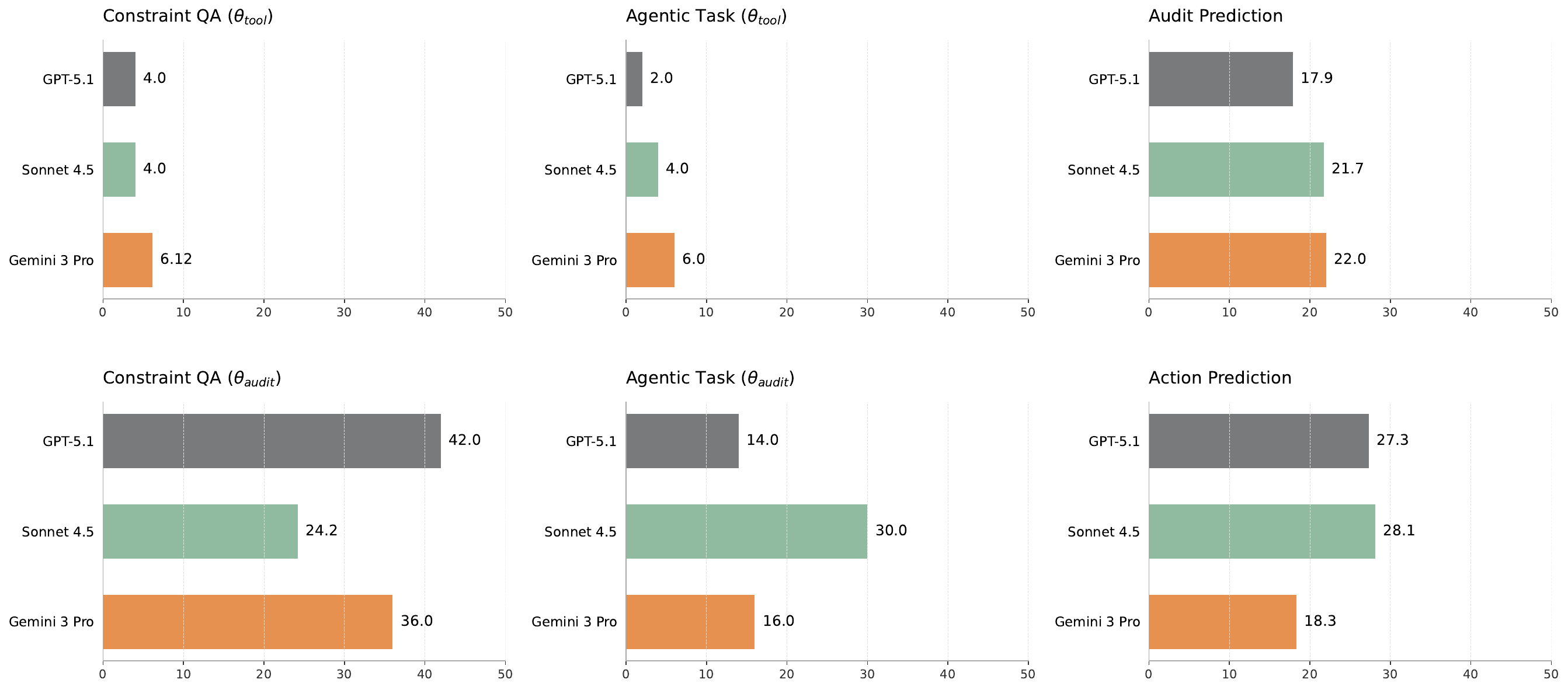}    
    \caption{Aggregated results for all task categories with frontier LLMs.}
    \label{fig:results_bar_plot}
\end{figure*}

\subsection{Results Overview}
WoW-bench evaluates several state-of-the-art LLMs including GPT-5.1, Gemini-3-pro, and Claude-Sonnet-4.5. Aggregated results are shown in~\Cref{fig:results_bar_plot}. 
The accuracy can significantly increase by at most 10x for constraint understanding tasks, indicating the necessity for a granular but tractable observation. 
As shown in~\Cref{tab:agentic}, while agents maintain reasonable task success rates (TSR) using standard tools, reliability under constraints (TSRUC) collapses near zero. GPT-5.1 achieves only 2\% TSRUC with $\mathcal{O}_{tool}$, which improves to 14\% with $\mathcal{O}_{audit}$. This confirms that without state visibility, agents blindly violate hidden constraints. For cost analysis, Claude's Opus-4.5 is the most expensive, while being on par with Gemini-3 for TSRUC.
For world modeling, LLMs have low accuracy (<30\% for both audit and action prediction), indicating a lack of dynamics modeling and domain understanding.

\section{Error Analysis}
Our evaluation reveals that agent failures in WoW are not merely stochastic errors in instruction following, but symptoms of fundamental gaps in how frontier LLMs represent state, model dynamics, and reason causally. We categorize these failures into three core deficiencies: The Representation Gap, The Dynamics Gap, and The Causal Gap. We provide detailed observations in~\Cref{appendix:additional_evaluation}.

\subsection{The Representation Gap: Lack of Symbolic Grounding}
A pervasive failure mode across all models is the inability to distinguish between semantic descriptions and symbolic identifiers. Models frequently conflate human-readable names with unique identifiers (e.g., predicting username vs. sys\_id), accounting for 73.5\% of errors. This suggests LLMs operate on surface-level tokens rather than a grounded entity abstraction, treating 'User X' as text rather than a node in a relational graph. In an enterprise database, "User X" is a node in a relational graph with specific properties (ID, Roles, Manager). LLMs, however, treat "User X" as a token in a context window. Without an explicit state abstraction, which is a structured mental representation of entities.

\subsection{The Dynamics Gap: Absence of Transition Models}
In Audit Prediction tasks, even the strongest models achieve near-zero accuracy in full state prediction. Models consistently under-predict the set of table audits, missing critical side effects (e.g., a Create Incident action silently updating the metric\_instance table). This indicates the lack of an explicit forward transition model ($P(s_{t+1} | s_t, a_t)$). Current agents rely on semantic similarity (e.g., creating an incident should update the incident table) rather than learning the environment's physics. In WoW, many transition rules are defined by hidden workflows. The failure to predict these side effects proves that LLMs are not simulating the world; they are merely hallucinating plausible completions based on general domain priors, which fail in environments governed by rigid, invisible business rules.

\subsection{The Causal Gap: Failure of Multi-Hop Reasoning}
The most severe failures occur in Agentic Tasks where constraints are violated by cascading effects (e.g., $Action \rightarrow Workflow A \rightarrow Workflow B \rightarrow Violation$). Agents successfully execute atomic actions but fail to detect that an early action (e.g., assigning a 4th asset) made a later action (e.g., assigning a restricted asset) impossible due to a clearance drop. This highlights a breakdown in causal rollout. To succeed, an agent must perform system 2 reasoning: projecting the state forward multiple steps to check for downstream violations. Our results show that LLMs are greedy planners, as they optimize for the immediate action's success while ignoring the long-term causal ripple effects. The attention decay observed in long-context constraint tasks further exacerbates this, as the causal link between a rule defined at $t=0$ and a violation at $t=10$ is lost in the context window.

\subsection{Implications: Toward Dynamics-Aware Agents}
The failures observed in WoW-bench demonstrate that scaling distinct capabilities such as context length, instruction following, or tool selection is insufficient for enterprise reliability. Instead, we argue for a paradigm shift toward dynamics-aware architectures that fundamentally change how agents perceive and interact with their environment.

Future agents must move beyond unstructured context windows to maintain explicit state abstractions. Our analysis shows that operating on surface-level text tokens leads to identity confusion and a loss of symbolic grounding. A reliable enterprise agent requires a structured mental representation that persistently tracks entity values, statuses, and relationships throughout the trajectory, independent of their textual mention in the conversation history.

Furthermore, reliability in these environments requires Model-Based Reinforcement Learning approaches where the agent learns a predictive model of the hidden workflows. Instead of purely reacting to tool outputs, a dynamics-aware agent should simulate the audit log before execution, using discrepancies between prediction and reality to update its internal model. This shift from reactive to predictive control allows the agent to anticipate compliance violations that are currently invisible to naive planners.

Finally, we posit that zero-shot performance is largely a fallacy in opaque enterprise systems. Successful agents must adopt active epistemic strategies rather than assuming a static world. This involves issuing probe actions to test hypotheses about workflow triggers and boundary conditions. By actively querying the environment to learn the physics of business rules, agents can transition from brittle instruction followers to robust systems capable of navigating the dynamics of real-world organizations.

\section{Conclusions}
WoW challenges the prevailing assumption that enterprise autonomy is merely a problem of better instruction following or broader tool use. Instead, we demonstrate that the fundamental bottleneck is dynamics understanding: the inability of current agents to model the invisible, cascading consequences of their actions in interconnected systems. WoW-bench serves as a standard testbed for this new frontier. It provides the first rigorous environment for measuring enterprise world modeling by isolating the capacity to predict hidden state transitions and trace multi-hop side effects. By shifting the evaluation focus from surface-level task completion to deep constraint reliability, we offer the community a necessary diagnostic tool to distinguish between agents that merely act and agents that understand.
Our findings motivate a transition away from purely reactive agent architectures. The observability gap we identified cannot be solved by prompt engineering alone; it demands a paradigm shift toward dynamics-aware systems that prioritize state grounding over simple text generation. Future work in enterprise AI should focus on the complex and hidden dynamics of real organizations. WoW provides the playground to build this next generation of systems, where agents that are not just compliant, but truly reliable.

\section{Limitations and Future Work}
WoW relies on experts to curate realistic tasks, constraints, and more importantly, workflows. The process requires domain knowledge for individual systems, making the expansion of the benchmark relatively expensive. Fortunately, such fixed cost is amortized over the long term and does not limit the growth or adoption of WoW. The diversity of the system dynamics only covers a subset of workflows in a single enterprise system. However, we believe such learned insights can transfer to other systems, as the core mechanics of hidden workflows are similar in most settings. WoW is intentionally designed as an evaluation benchmark rather than a training environment. Our goal is to diagnose limitations of current agents under realistic, partially observable system dynamics; demonstrating improved performance through training-time interaction is an important direction for future work and is enabled by WoW’s fully interactive design.

\bibliography{main}
\bibliographystyle{icml2026}

\newpage

\appendix
\section{Action and Observation Examples}
\label{appendix:tool}

The main method of interaction is MCP tools for ServiceNow's API endpoints. We adopt an existing ServiceNow MCP server~\footnote{\href{https://github.com/echelon-ai-labs/servicenow-mcp}{Echelon Lab MCP}} and extended it to a total of 108 tools for more functionalities. The actions include four types of database and API operations known as CRUD: create, read, update, and delete.
We choose MCP tools instead of directly calling APIs to constrain the action space and provide ease of use for researchers without domain knowledge. However, the action space is still combinatorially large, as the inputs to the parameters are free-form text instead of enumeration. To ensure the validity of generated actions, we provide constraints of mandatory input and the format of each input in the tool specification, and parse the structured output. In our analysis, frontier LLMs consistently output valid actions under these constraints.

\begin{promptbox}{Example of Action}
\begin{lstlisting}[basicstyle=\ttfamily\small, breaklines=true]
{
  "action": {
    "tool_name": "create_incident",
    "parameters": {
      "short_description": "Server outage in production environment",
      "description": "The server is down and we are unable to access the production environment",
      "impact": "1",
      "urgency": "1"
    }
  }
}
\end{lstlisting}
\end{promptbox}

\begin{promptbox}{Example of Observation as StateDiff}
\begin{lstlisting}[basicstyle=\ttfamily\small, breaklines=true]
{
  "sysauditrecord": [
    {
      "fieldname": "calculation_complete",
      "newvalue": "0",
      "tablename": "metric_instance",
      "oldvalue": ""
    },
    {
      "fieldname": "calculation_complete",
      "newvalue": "1",
      "tablename": "metric_instance",
      "oldvalue": ""
    },
    {
      "fieldname": "table",
      "newvalue": "incident",
      "tablename": "metric_instance",
      "oldvalue": ""
    },
    {
      "fieldname": "value",
      "newvalue": "false",
      "tablename": "metric_instance",
      "oldvalue": ""
    },
    {
      "fieldname": "field_value",
      "newvalue": "1",
      "tablename": "metric_instance",
      "oldvalue": ""
    }
    /* ... continue all entries ... */
  ],
  "additional_information": {
    "num_audits": 88,
    "num_modified_entries": 10,
    "num_deleted_entries": 0,
    "num_created_entries": 78,
    "operation_type": ["put"],
    "tables_modified": [
      "sla_breakdown_by_assignment",
      "metric_instance",
      "task_sla",
      "incident"
    ]
  }
}
\end{lstlisting}
\end{promptbox}

\begin{promptbox}{Tool Specification Example for Update Incident}
\begin{lstlisting}[style=python]
from pydantic import BaseModel, Field
from typing import Optional
class UpdateIncidentParams(BaseModel):
    """Parameters for updating an incident."""

    incident_id: str = Field(..., description="Incident ID or sys_id")
    short_description: Optional[str] = Field(None, description="Short description of the incident")
    description: Optional[str] = Field(None, description="Detailed description of the incident")
    state: Optional[str] = Field(None, description="State of the incident")
    category: Optional[str] = Field(None, description="Category of the incident")
    subcategory: Optional[str] = Field(None, description="Subcategory of the incident")
    priority: Optional[str] = Field(None, description="Priority of the incident")
    impact: Optional[str] = Field(None, description="Impact of the incident")
    urgency: Optional[str] = Field(None, description="Urgency of the incident")
    assigned_to: Optional[str] = Field(None, description="User assigned to the incident")
    assignment_group: Optional[str] = Field(None, description="Group assigned to the incident")
    work_notes: Optional[str] = Field(None, description="Work notes to add to the incident")
    close_notes: Optional[str] = Field(None, description="Close notes to add to the incident")
    close_code: Optional[str] = Field(None, description="Close code for the incident")
\end{lstlisting}
\end{promptbox}

\section{Details of the Benchmark Suite}
\label{appendix:dataset}

\subsection{Annotation Process}

\textbf{Constraint Understanding Tasks} To create realistic and task-specific constraints, we ask domain experts to manually create 10 realistic constraints. An example of the constraint tackling the user and incident management is \emph{A user cannot be assigned more than 3 active incidents at a time}. Another constraint focusing on knowledge base is \emph{Flagged articles should not be published}. One could argue that such constraints can be forced with rules embedded in the system. However, they can be expensive to set up, ad-hoc, time-dependent, and may not necessarily apply to all use cases in an enterprise. To create trajectories that violate these constraints, we ask domain experts to carefully examine the effects of hidden business rules and workflows, then design the actions to have cascading effects on purpose. We curate 10 templates for constraint violation trajectories and further perturb them and obtain 50 trajectories in total. Furthermore, to study the difference in $\Omega_{tool}$ and $\Omega_{audit}$, we evaluate separately when agents take $\mathcal{O}_{tool}$ and $\mathcal{O}_{audit}$ as inputs. 

\textbf{Agentic Task Completion} There are 10 task templates that have been crafted taking inspiration from the 10 original constraint understanding trajectories. The trajectories have been converted into task descriptions by an LLM and verified by human experts. The task description outlines the sequence of actions that the LLM agents must perform. 
The agents are prompted to follow the chronology outlined in the task description closely instead of skipping forward to the effective end result. For example, if an asset was initially assigned to user1 and then transferred to user2, the agent can't simply assign the asset to user2 since it will be transferred anyway. In actual enterprise settings, these operations could happen over different time horizons and there could be many constraints that might be violated in such operations, which is what we wanted to study. 

On average, it requires 13 actions to complete a task.
There are a total of 10 task templates with 5 permutations each (50 total tasks). Each task template focuses on one specific constraint violation. Because of overlap of constraint violations, 2 out of 10 task templates have multiple constraints that can be violated. These 10 templates cover all the 10 constraints that had been set up before. 

\textbf{Action and Audit Prediction} We use three main methods to create the trajectories for action and audit prediction tasks. Note that the same trajectory is used for both action and audit prediction. For action prediction, we use the audits as input and prompt LLMs to generate all actions at once. For audit prediction, we use $K$ actions as input and prompt the LLMs to generate $K$ audits, where $K=(1,2,3,4,5)$. To create trajectories, all three methods use heuristics to generate actions, and execute the actions in the WoW environment to obtain the audits. The first one is manual creation, where human experts are asked to utilize as many tools as possible to create diverse trajectories. These trajectories are high-quality and can represent real task trajectories. The second approach is graph-based sampling, which aims to remove human effort of manual creation. We add a tool-dependency graph based sampling for trajectories. This involves building a graph of the tools, where each node is a tool, and the edge between tools indicate which tools' outputs can plug into another tool's input. The sampling process involves randomly picking a tool from the total tool list, and then backtracking all the way to the root node (A node without any inbound edges). The details of the sampling approach can be found in Appendix~\ref{appendix:dataset}. The benefits of the graph-based approach is : 1) WoW acts as a true environment which does not rely on fixed trajectories as a simple benchmark. It becomes an environment which provides the mechanism to sample trajectories and becomes a learning environment that can be trained on. 2) Most of the workflows need a series of connected tool calls to reach states that completely random disjoint tool calls would not be able to explore.
The last approach is random sampling, where we provide the full suite of MCP tools to an LLM (i.e. GPT-4O) and prompt it to randomly perform actions up to $K$ steps. Note that this approach should not be the only method to scale up the number of trajectories, as diversity can become limited for exploring different states, even with carefully designed diverse prompts.

In our analysis, many system tables are also populated, and some actions can introduce hundreds of corresponding changes in the system tables. However, such system tables usually contain specific records regarding the backend setup, configurations, etc., which are not semantically useful for understanding the state needed to solve common enterprise tasks. We decided to remove them to keep the observation tractable. Such system tables could potentially contribute to understanding the environmental dynamics in more complex tasks, but they are currently outside the scope of this work.

\begin{figure*}[htbp]
    \centering
    \includegraphics[width=0.9\textwidth]{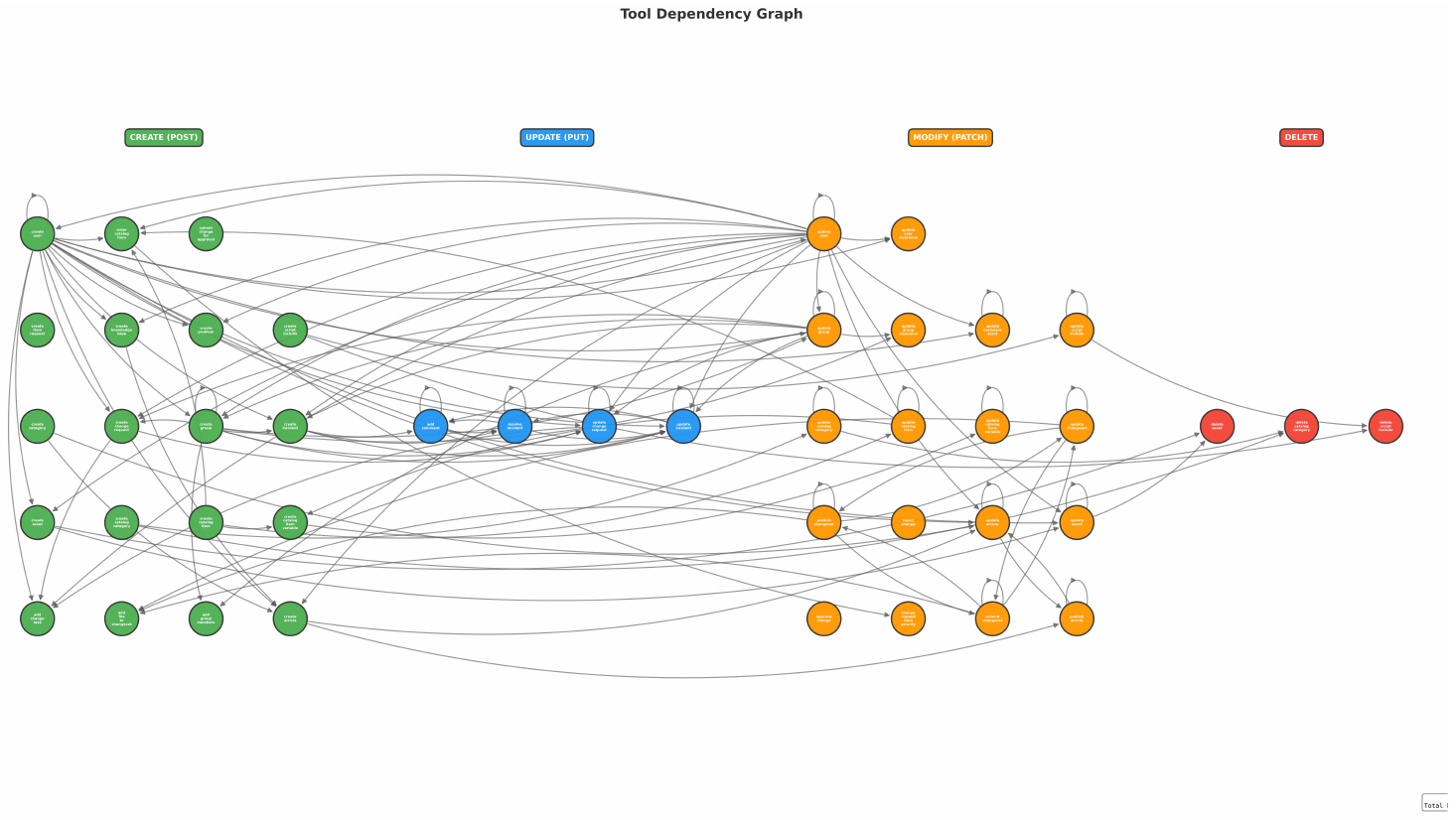}
    
    
    \caption{Tool-based dependency graph}
    \label{fig:tool_graph}
\end{figure*}

\textbf{Tool-Graph Based Trajectory Sampling}. We propose a trajectory sampling method that removes the previous human effort of manually running MCP tools and capturing the trajectories for evaluation. In the updated approach, we add a tool-dependency graph~\Cref{fig:tool_graph}-based sampling method for trajectories. This involves building a graph of the tools, where each node is a tool, and the edges between tools indicate which tools' outputs can plug into another tool's input. The sampling process involves randomly picking a tool from the total tool list, and then backtracking all the way to the root node (a node without any inbound edges). The backtracking is performed only on the mandatory inputs of the tool. Thereafter, based on a uniform probability distribution, either a random tool is sampled and this process is repeated, or some of the optional reference inputs of the initially sampled tool are picked and the backtracking is repeated. This continues until we reach a maximum threshold for trajectory length. The input and output links are determined based on the input parameters of the tool, since these input parameters directly plug into the payload for a specific table in ServiceNow. A custom scripted REST API is created that checks if a given field in a table is a reference to another table. This is then linked to the output of the tool by looking at the final POST/PUT/PATCH request of the MCP tool, since the output will be a record from the table to which it made an edit.

The benefit of doing this, apart from removing the human-in-the-loop, is: 1) WoW acts as a true environment that does not rely on fixed trajectories as a simple benchmark. It becomes an environment that provides the mechanism to sample trajectories and becomes a learning environment that can be trained on. 2) Most workflows need a series of connected tool calls to reach states that completely random, disjoint tool calls would not be able to explore. A connected tool call sequence allows better state exploration and a higher chance of exploring the side effects of workflows that trigger only when repeated tool calls are made regarding the same database entity. 

\begin{promptbox}{List of All Tables Shown in the Audits}
\begin{lstlisting}[basicstyle=\ttfamily\small, breaklines=true]
alm_asset
alm_hardware
asmt_assessable_record
asset_job_log
change_request
cmn_notif_device
cmn_notif_message
ecc_queue
fm_expense_line
fx_currency_instance
fx_price
incident
item_option_new
kb_knowledge
kb_knowledge_base
kb_uc_can_contribute_mtom
kb_uc_can_read_mtom
kb_uc_cannot_contribute_mtom
kb_uc_cannot_read_mtom
metric_instance
pc_hardware_cat_item
problem
pwd_enrollment_snapshot
sc_cart
sc_cart_item
sc_cat_item
sc_cat_item_catalog
sc_cat_item_category
sc_category
sc_req_item
sc_request
sc_task
sla_breakdown_by_assignment
stage_state
sys_script_include
sys_update_set
sys_user
sys_user_grmember
sys_user_group
sys_user_has_role
task_sla
ua_upload_log
ua_upload_log_details
v_user_session
var__m_var_dictionary_null
var__m_wf_activity_variable_
wf_activity
wf_condition
wf_context
wf_executing
wf_history
wf_log
wf_transition_history
wf_workflow
wf_workflow_binding
\end{lstlisting}
\end{promptbox}

\begin{promptbox}{List of All Constraints}
1. Moving a US-based asset to a different country requires approval first. \\
2. Flagged articles should not be published. \\
3. Users should not be assigned the same role as their managers. \\
4. A user should not be assigned more than 3 active incidents at a time. \\
5. A problem should always have an assigned user. \\
6. Assets cannot be transferred to users whose clearance\_level is below the asset's required\_clearance\_level. \\
7. Cannot transfer a user from one group to another if they would lose clearance and be forced to surrender assets. \\
8. High-priority requests cannot be rejected on the grounds of an inactive user. \\
9. Priority 1 task derivative records cannot be canceled, closed, or rejected. They can only be resolved. \\
10. Assets with a value greater than 10000\$ cannot be transferred to users who already have more than 2 active assets.
\end{promptbox}

\section{Prompt}
\label{appendix:prompt}

\begin{promptbox}{Partial State Prediction Prompt}

You are an IT Service Management (ITSM) expert. More specifically, you are a ServiceNow expert who is well familiar with the ServiceNow Developer instance and is aware of all the underlying changes that happen in different ServiceNow tables after making
changes in one of them. You will serve as an enterprise world model for state prediction.

You will be given the input including the initial state, a sequence of actions (which in this case are MCP tool calls), the schema of all the tables, and a description of the user task. As a world model, your main task is to understand how the action will change the underlying state of the world. The state includes two major parts:

1. The table audits have the following JSON structure with four keys: \texttt{fieldname}, \texttt{newvalue}, \texttt{tablename}, and \texttt{oldvalue}. The following is an example of an update operation for the column \texttt{short\_description} in table \texttt{sc\_req\_item}. There will usually be multiple table audits associated with one action.

{\small
\begin{verbatim}
{
  'fieldname': 'short_description',
  'newvalue': 'Updated sc_req_item 
  from API',
  'tablename': 'sc_req_item',
  'oldvalue': 'Test sc_req_item 
  from API',
}
\end{verbatim}
}

2. An \texttt{AdditionalInformation} section for each state that specifies the total number of audits (\texttt{num\_audits}), updated columns (\texttt{num\_modified\_entries}), deleted columns (\texttt{num\_deleted\_entries}), created entries (\texttt{num\_created\_entries}), operation type (a list of operation types that the action performs, can only be [get, put, post, delete]), and all table names that appear in the table audits (\texttt{tables\_modified}).

Your output, which is the \texttt{StateDiff}, should strictly follow this schema:

{\small
\begin{verbatim}
{
  "sysauditrecord": [
    {
      "fieldname": "field_that_changed",
      "tablename": "table_name", 
      "oldvalue": "previous_value",
      "newvalue": "new_value"
    }
  ],
  "additional_information": {
    "num_audits": 1,
    "num_modified_entries": 0,
    "num_deleted_entries": 0,
    "num_created_entries": 1,
    "operation_type": ["put"],
    "tables_modified": ["table_name"]
  }
}
\end{verbatim}
}

\noindent
\textbf{Table Schemas:} These are the table schemas for some relevant tables. The keys in the dictionary are the table names, and the entries correspond to the different columns for each table. The column name is represented by \texttt{"element"}, \texttt{"mandatory"} indicates whether it is a mandatory field, \texttt{"reference"} indicates whether the column value references a different table, and \texttt{"internal\_type"} indicates the data type for the column value. \texttt{"default\_value"} specifies the default value of the field if not directly specified.

\begin{verbatim}
{TABLE_SCHEMA}
\end{verbatim}

\noindent
\textbf{MCP Tools:} The following is a description of all the MCP Tools (these are the set of total actions, some of which will show up in the trajectory given below).

\begin{verbatim}
{MCP_TOOLS}
\end{verbatim}

Some examples of actions and their corresponding states are:
\begin{verbatim}
{EXAMPLES}
\end{verbatim}
\end{promptbox}

\begin{promptbox}{Action Prediction Prompt}

You will serve as an enterprise world model for state prediction. Specifically, the system environment you are operating in is the ServiceNow developer instance.

You will be given the input including the initial state, a sequence of state changes, the schema of all the tables, and a description of the user task. As a world model, your main task is to understand what actions will change the underlying state of the world and predict them. The state includes two major parts:

1. The table audits have the following JSON structure with four keys: \texttt{fieldname}, \texttt{newvalue}, \texttt{tablename}, and \texttt{oldvalue}. The following is an example of an update operation for the column \texttt{short\_description} in table \texttt{sc\_req\_item}. There will usually be multiple table audits associated with one action.

{\small
\begin{verbatim}
{
  'fieldname': 'short_description',
  'newvalue': 'Updated sc_req_item from 
  API',
  'tablename': 'sc_req_item',
  'oldvalue': 'Test sc_req_item from 
  API',
}
\end{verbatim}
}

2. An \texttt{AdditionalInformation} section for each state that specifies the total number of audits (\texttt{num\_audits}), updated columns (\texttt{num\_modified\_entries}), deleted columns (\texttt{num\_deleted\_entries}), created entries (\texttt{num\_created\_entries}), operation type (a list of operation types that the action performs, can only be [get, put, post, delete]), and all table names that appear in the table audits (\texttt{tables\_modified}).

There will be multiple state changes in a sequence, and your primary job is to identify the actions leading to each state change.

Your output, which is a sequence of MCP tools (actions), should strictly follow this schema:

{\small
\begin{verbatim}
[
  {
    "actions": {
      "tool_name": "tool_1",
      "parameters": {
        "parameter_1": "value",
        "parameter_2": "value"
      }
    }
  },
  {
    "actions": {
      "tool_name": "tool_2",
      "parameters": {
        "parameter_1": "value",
        "parameter_2": "value"
      }
    }
  }
]
\end{verbatim}
}

\noindent
\textbf{Table Schema:} The following describes the tables and columns present in the environment.

{\small
\begin{verbatim}
{TABLE_SCHEMA}
\end{verbatim}
}

\noindent
\textbf{MCP Tools:} These are the available actions that the model can invoke during prediction.

\begin{verbatim}
{MCP_TOOLS}
\end{verbatim}

Some examples of actions and their corresponding states are:

\begin{verbatim}
{EXAMPLES}
\end{verbatim}

\end{promptbox}

\section{Additional Evaluation}
\label{appendix:additional_evaluation}
Table~\ref{tab:full_state_table} shows the full state prediction accuracy, which is calculated by exact matching of all table audits (including both field and values). This is a much more difficult task, since the LLM has to understand exactly how the system operates. As a result, the LLM predicts the state transition with 0\% full accuracy.

The errors in state predictions include wrong predictions (not understanding the consequence of an action or predicting the wrong data type '0' v.s. 'False'), false positives (predicting side-effects that do not happen), and false negatives (missing critical changes). For example, when applying Gemini-2.5-pro on the composite4 task for state prediction, the audit on the fm\_expense\_line.type field is missing, while the audit on the alm\_asset.display\_name field is a false positive. 
For action predictions, most errors are parameter-related. An interesting observation is that GPT-5 performs worse than O3, some of its mistakes are obfuscating create vs. read operations, for example, predicting get\_user instead of create\_user operation. 

\subsection{Detailed Error Modes}
\subsubsection{Constraint Understanding} 
When given $\mathcal{O}_{tool}$, agents can only reason whether constraints are violated by examining actions and tool responses. Since they do not expose the full database state change, LLMs often fail to identify violations.
When $\mathcal{O}_{audit}$ is used, LLMs tend to be very accurate for trajectories containing only a single constraint violation. However, compositional constraint violation tasks pose challenges. This could be due to the attention decay problem in long context length, since table audits do increase the input length significantly. On average, the trajectory length for single constraint violation is 53K tokens, where it increases to 109K tokens for compositional constraints.

\subsubsection{Agentic Task Completion} 
We identify several common issues for all models. The first one is the inability to fully track the symbolic state within the audit logs. Even if the full audit logs are available to agents, they cannot keep the entities, such as users and incidents, fully up to date due to the number of actions and workflow triggers. The second observation is closely related, which is the lack of multi-hop reasoning capability, especially in tasks that involve long trajectories and chains of data flow. When the target entities referenced in the MCP tools go through multiple changes causing workflow triggers, other tables can be modified and task completion can be reverted. However, the task is still perceived to be completed by the agents. Both the aforementioned errors are commonly detected. There are model-specific errors for the Claude family, where structured outputs sometimes fail to be parsed. Moreover, they also can misinterpret the tool specification and misuse tools, which was not observed for other frontier models. 
For the tasks that are inherently impossible to complete due to constraints, none of the models are able to correctly identify them. This is especially alerting, as agents are specifically prompted to report such cases, yet fail to do so. On average, LLMs take 11 to 13 steps to complete the tasks depending on the models, which is fewer than the oracle trajectory length. Agents generally do not attempt to reason on or revert constraint violation.

\subsubsection{Action Prediction}
From our experiments, GPT-5.1 and Gemini-3-Pro struggle more ($60-70\%$ of total errors) on predicting the correct parameters while Sonnet-4.5 fails more frequently ($60\%$) on selecting the right MCP tools. For tool name prediction, the incorrect tool name predictions from Gemini-3-Pro are more evenly distributed. However, GPT-5.1 tend to predict more actions wrongfully as \textbf{list\_user}, and Sonnet-4.5 frequently predicts other actions incorrectly as \textbf{search\_any\_table}. For tool parameter prediction, all three models struggle on sys\_id fields, for example, the user\_id and incident\_id. For example, these models frequently predicts username instead of sys\_id for users (accounting for 73.5\% of user\_id failures) and predicts incident number instead of sys\_id for incidents (accounting for 91.2\% of incident\_id failures), which shows that LLMs could have difficulties in differentiating similar field names for MCP tools or they may blindly opt for the more human readable format, e.g.,  username instead of the UUIDs. 




\subsubsection{Audit Prediction}
Overall, the GPT-5.1 and Sonnet-4.5 model tend to under-predict audits significantly while Gemini-3-Pro predicts similar number of audits compared to the ground truth. Among all actions, all three models perform well ($>50\%$) IoU on user management actions but perform poorly on knowledge base and catalog item management actions. 
Moreover, we observe that all three models' state prediction accuracy first increases from $0\%$ to ($10\%-20\%$) when $k$ increases from $1$ to $4$ before dropping at $k=4$. Usually, the instinct is that when $k$ increases, the error will accumulate resulting in lower accuracy in state prediction. However, this environment is partially observable and at the first step $k=1$ no initial state is provided, which could explain why all three models perform poorly with 0\% full accuracy at the first step (\ref{tab:full_state_table}). Overall, this experiment indicates that LLM's capability could be significantly affected by the observability of the environment.

\begin{figure}[ht]
    \centering
    \includegraphics[width=0.9\columnwidth]{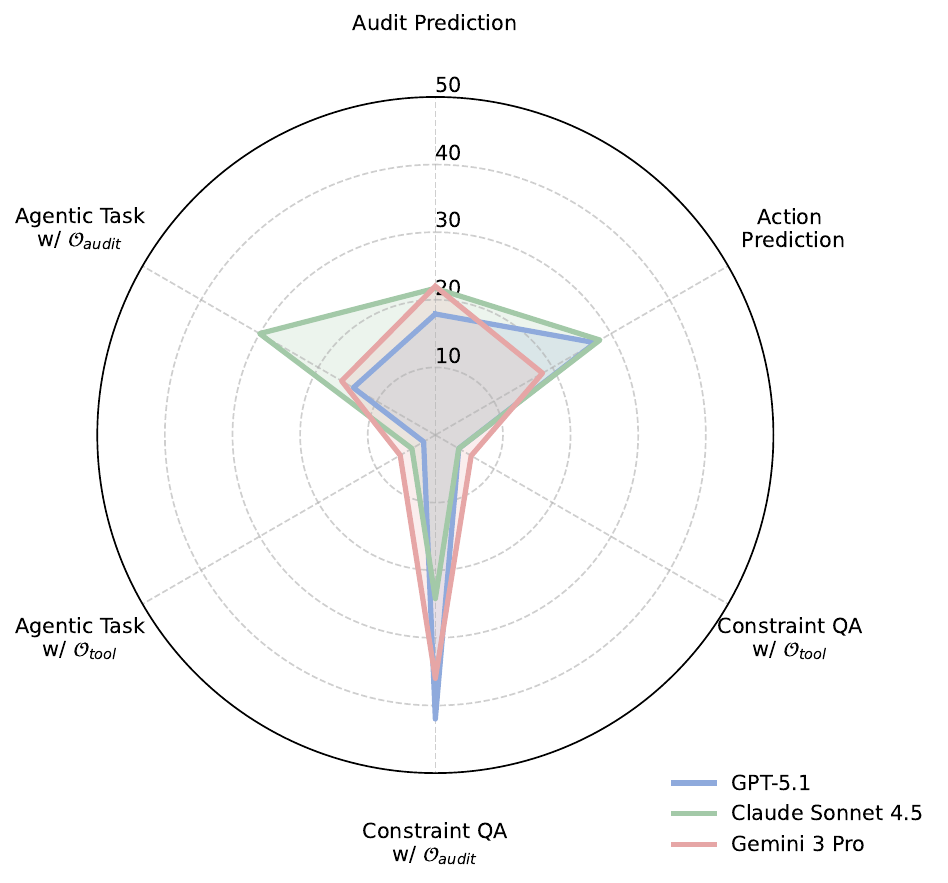}    
    \caption{Performance of frontier LLMs across task categories. Constraint understanding and agentic tasks are evaluated using two observation modes: tool response $\mathcal{O}_{tool}$ and table audits $\mathcal{O}_{audit}$.}
    \label{fig:wow_radar}
\end{figure}

\begin{figure*}[ht]
    \centering
    \includegraphics[width=0.9\textwidth]{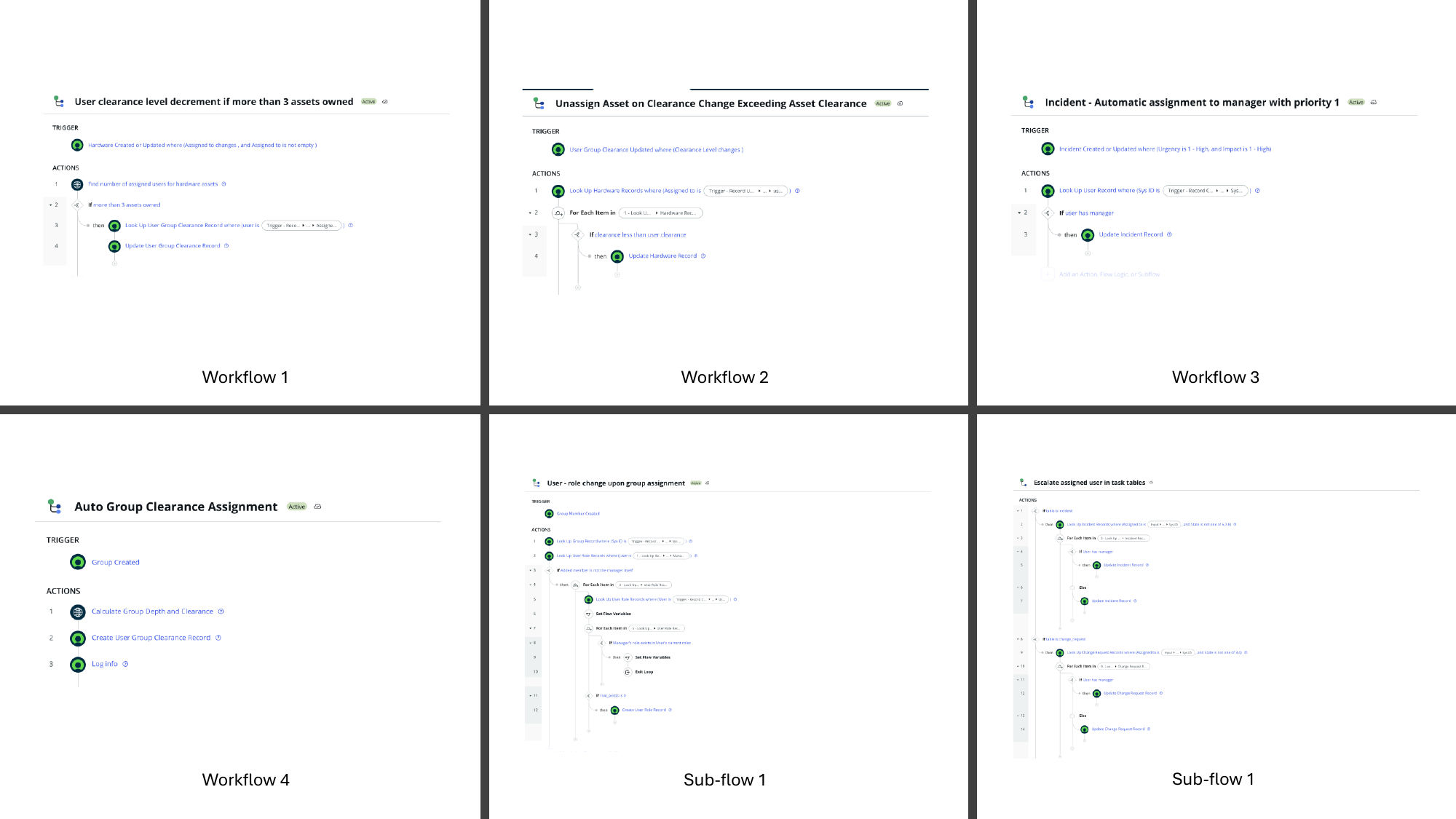}    
    \caption{Examples of workflows in WoW.}
    \label{fig:workflow_screenshot}
\end{figure*}

\begin{table}[t]
\centering
\caption{Constraint understanding accuracy for $\mathcal{O}_{audit}$ and $\mathcal{O}_{tool}$.}
\begin{tabularx}{\columnwidth}{c Y Y}
\toprule
\textbf{Task} & Acc. w/ $\mathcal{O}_{audit}$ & Acc. w/ $\mathcal{O}_{tool}$\\ \midrule
OpenAI O3 & 30.6\% & 4.0\% \\
GPT-5.1 & 42.0\% & 4.0\% \\
Gemini-2.5-Pro & 36.0\% & 6.12\% \\
Sonnet-4.5 & 24.2\% & 4.0\% \\ \bottomrule
\end{tabularx}
\label{tab:constraint}
\end{table}

\begin{table}[ht]
\centering
\caption{Action prediction results, where tool name and parameter accuracies are both reported. $\text{ACC}_{\text{Type}}$ is the tool name accuracy, and $\text{ACC}_{\text{Full}}$ is the full accuracy.}
\begin{tabularx}{\columnwidth}{c Y Y}
\toprule
\textbf{Model} & $\text{ACC}_{\text{Type}}$ & $\text{ACC}_{\text{Full}}$  \\ \midrule
GPT-5.1 & 77.8\% & 27.3\% \\
Gemini-3-Pro & 79.0\% & 18.3\% \\
Sonnet-4.5 & 81.7\% & 28.1\% \\ \bottomrule
\end{tabularx}
\label{tab:action_pred}
\end{table}

\begin{table}[ht]
\centering
\caption{Audit prediction IOU with K-step rollout.}
\begin{tabularx}{\columnwidth}{c Y Y Y Y Y Y}
\toprule
\textbf{Model} & K=1& K=2& K=3& K=4& K=5 & Avg.  \\ \midrule
GPT-5.1 & 22\% & 16\% & 21\% & 15\% & 15\% & 18\% \\
Gemini-3-Pro & 32\% & 20\% & 23\% & 19\% & 13\% & 22\%\\
Sonnet-4.5 & 31\% & 19\% & 23\% & 20\% & 13\% & 22\% \\ \bottomrule
\end{tabularx}
\label{tab:state_pred}
\end{table}

\begin{table}[ht]
\centering
\caption{Audit prediction accuracy with K-step rollout, where the accuracy is calculated from exact matching of field and value of all table audits.}
\begin{tabular}{c|ccc}
\toprule
\textbf{Model} & \textbf{GPT-5.1} & \textbf{Gemini-3-Pro} & \textbf{Sonnet-4.5} \\ \midrule
K=1 & 0\% & 0\% & 0\%  \\ 
K=2 & 7.1\% & 13.2\% & 10.7\%  \\ 
K=3 & 1.9\% & 4\% & 3.8\% \\ 
K=4 & 10.4\% & 17.8\% & 18.8\% \\ 
K=5 & 8.9\% & 9.5\% & 8.9\% \\ \midrule
Avg. & 5.4\% & 8.6\% & 8.1\%  \\ \bottomrule
\end{tabular}
\label{tab:full_state_table}
\end{table}

\begin{table}[ht]
\centering
\caption{Evaluation for single-trajectory constraint understanding.}
\begin{tabular}{ccccc}
\toprule
\textbf{Model} & \textbf{O3} & \textbf{GPT-5} & \textbf{Gemini-2.5-Pro} & \textbf{Sonnet-4.5} \\ \midrule
$\mathcal{O}_{audit}$ & 80.0\% & 80.0\% & 70.0\% & 80.0\% \\
$\mathcal{O}_{tool}$      & 10.0\% & 20.0\% & 20.0\% & 10.0\% \\ \bottomrule
\end{tabular}
\label{tab:constraint_single}
\end{table}

\end{document}